# Multi-view Fake News Detection Model Based on Dynamic Hypergraph

**Rongping Ye, Xiaobing Pei**

Huazhong University of Science and Technology

m202276628@hust.edu.cn, xiaobingp@hust.edu.cn

## Abstract

With the rapid development of online social networks and the inadequacies in content moderation mechanisms, the detection of fake news has emerged as a pressing concern for the public. Various methods have been proposed for fake news detection, including text-based approaches as well as a series of graph-based approaches. However, the deceptive nature of fake news renders text-based approaches less effective. Propagation tree-based methods focus on the propagation process of individual news, capturing pairwise relationships but lacking the capability to capture high-order complex relationships. Large heterogeneous graph-based approaches necessitate the incorporation of substantial additional information beyond news text and user data, while hypergraph-based approaches rely on predefined hypergraph structures. To tackle these issues, we propose a novel dynamic hypergraph-based multi-view fake news detection model (DHy-MFND) that learns news embeddings across three distinct views: text-level, propagation tree-level, and hypergraph-level. By employing hypergraph structures to model complex high-order relationships among multiple news pieces and introducing dynamic hypergraph structure learning, we optimize predefined hypergraph structures while learning news embeddings. Additionally, we introduce contrastive learning to capture authenticity-relevant embeddings across different views. Extensive experiments on two benchmark datasets demonstrate the effectiveness of our proposed DHy-MFND compared with a broad range of competing baselines.

## Introduction

As social media experiences rapid growth, numerous social platforms have emerged as primary sources for people to gather information in their everyday lives. However, due to the ease of information propagation on social media and the inadequacies in content moderation mechanisms, the propagation of fake news has become a significant issue. To combat the negative effects associated with the spread of fake news, various methods for fake news detection have been proposed by researchers in recent years.

Initially, numerous methods leveraging NLP models for news text mining were proposed to detect fake news (Kaliyar et al. 2021; Rao et al. 2021). However, the deceptive nature of fake news hampers the effectiveness of text-based approaches (Gong et al. 2023). Consequently, considering how news spreads across social networks, including interac-

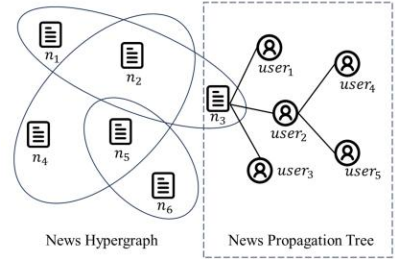

Figure 1: News hypergraph and news propagation Tree.

-tions like comments and shares, researchers construct news propagation trees. Several graph neural network methods have been proposed to learn these propagation trees (Bian et al. 2020; Lu et al. 2020; Dou et al. 2021). However, these approaches are constrained to individual news propagation processes, prompting the exploration of more complex structures and methods. Methods based on large heterogeneous graph further integrate additional information across multiple news propagation processes (Nguyen et al. 2021; Cui et al. 2022). Hypergraph-based methods model complex high-order relationships among multiple news pieces as hypergraph structures (Jeong et al. 2022).

However, there are certain shortcomings in existing methods. Firstly, the majority of propagation tree-based methods are constrained by information limited to individual news propagation processes, neglecting potential connections with other relevant news, thus restricting their performance in fake news detection. Secondly, large heterogeneous graph-based methods require a considerable amount of additional information beyond news content and user profiles. While the additional data contains useful information, acquiring and processing it comes with expensive costs and the potential introduction of noise. Lastly, hypergraph-based methods employ predefined hypergraph structures, which do not guarantee the quality of these structures.

To tackle the aforementioned drawbacks, we propose a novel method named **D**ynamic **Hy**pergraph-based **M**ulti-view **F**ake **N**ews **D**etection model (DHy-MFND). To begin with, we learn news embeddings from three distinct views: the text-level view, the news propagation tree-level view, and the news hypergraph-level view. In the text-level view, we utilize pre-trained NLP models to encode news text data and associated user profiles, resulting in news text embeddings and user embeddings. These embeddings act as initial node features for the other two views. In the propagation tree-level view, we employ GNNs to capture the news propagation features. In the hypergraph-level view, we employ HGNNs to capture high-order relationships among news pieces. Then, we introduce dynamic hypergraph structure learning to optimize the predefined hypergraph structure by leveraging the similarity between news node embeddings and hyperedge embeddings. Additionally, we incorporate contrastive learning between news embeddings extracted from the propagation tree-level view and the hypergraph-level view to emphasize capturing authenticity-relevant embeddings. Here, pairs of news sharing the same authenticity are considered as positive pairs, while pairs with differing authenticity are considered as negative pairs. In the end, we use a self-attention mechanism to merge the news embeddings learned from the three views into final news embeddings. Our main contributions are summarized as follows:

- We propose a novel model named DHy-MFND for fake news detection, which involves learning news embeddings from three distinct views. It utilizes pre-trained NLP models, GNNs, and HGNNs to capture embeddings from each view. To obtain authenticity-relevant embeddings, a contrastive learning mechanism is incorporated between the embeddings from the propagation tree-level view and the hypergraph-level view. Subsequently, a multi-view embedding fusion process is conducted to derive the ultimate news embeddings.
- To explore high-order relationships among fake news, DHy-MFND introduces a dynamic hypergraph convolution block that optimizes the hypergraph structure while learning news embeddings.
- Comprehensive experiments are conducted on two well-known fake news detection datasets and the results demonstrate that DHy-MFND consistently outperforms various competing baselines.

## Related Work

### Hypergraph Learning

As an extension of graphs with defined hyperedges, a hypergraph consists of nodes and hyperedges, where a hyperedge can contain two or more nodes (Gao et al. 2021; Yang et al. 2022). Due to this characteristic of hyperedge, hypergraphs possess an advantage over general graphs in modeling non-pairwise relationships. The rise of complex structured data in diverse applications like recommendation systems (Wu et al. 2024), community detection (Ruggeri et al. 2023), and link prediction (Wang et al. 2023) has led to growing interest in hypergraph learning. Efforts have been made to extend neural networks designed for general graphs to hypergraphs, with HGNN (Feng et al. 2019) pioneering the extension of spatial methods to hypergraphs, thus enabling the capture of high-order relationships. However, the majority of these works concentrate on static hypergraph structures, where model performance is dependent on the quality of the hypergraph structure.

To address this issue, recent effort has been undertaken to explore learning from dynamic hypergraph. DHGNN (Jiang et al. 2019) dynamically constructs hypergraphs using the KNN algorithm and k-Means clustering algorithm before conducting hypergraph convolution. HERALD (Zhang et al. 2021) optimizes the hypergraph Laplacian matrix. HSL (Cai et al. 2022) employs node and hyperedge sampling, integrating contrastive learning within hyperedges to learn sparse hypergraph structures.

### Fake News Detection

With the rapid expansion of online social networks in recent years, fake news detection has emerged as a pressing concern, leading to the proposal of numerous news text mining methods to address this issue (Kaliyar et al. 2021; Rao et al. 2021). FinerFact (Jin et al. 2022) extracts claims from news content and evidence from comments, constructing a fully connected graph using GAT encoding. Subsequently, efforts have been made to utilize more information for fake news detection, leading to the emergence of news propagation tree-based methods. These methods focus on the propagation process of news across social networks, considering interactions like comments and reposts to construct a news propagation tree. Various GNN-based approaches have been proposed to learn from the news propagation tree. BiGCN (Bian et al. 2020) encodes the news propagation tree by employing two graphs in a top-down and bottom-up manner. Subsequently, researchers have attempted to incorporate more additional information. GCAN (Lu et al. 2020) gathers statistical information and constructs a user fully connected graph, UPFD (Dou et al. 2021) utilizes user preferences based on historical data, and GCNFN (Monti et al. 2019) uses a two-layer GCN to model a propagation tree with the fusion of heterogeneous data. To address potential noise and credibility issues within propagation trees, RDEA (He et al. 2021) and GACL (Sun et al. 2022) integrate contrastive learning into graph-based fake news detection.

Complex structures and methods have been introduced to leverage the latent relationships among multiple news pieces. To capture relationships between different news, UniPF (Wei et al. 2022) leverages the similarity between propagation tree structures and introduces information interactions between these trees. In order to utilize more information, large heterogeneous graph-based methods have been proposed, further incorporating relationships such as information sources across multiple news propagation trees (Nguyen et al. 2021; Cui et al. 2022). SureFact (Yang et al. 2022) constructs a heterogeneous graph using news, users, posts, and keywords, using reinforcement learning to select important subgraphs for fake news detection. Hypergraph-based methods model multiple news pieces as hypergraph structures, utilizing the similarities between news articles (such as topics, styles, content, etc.). HGFND (Jeong et al. 2022) constructs hypergraph structures based on the similarities between news.

## Preliminaries

### Notations and Definition

In the problem of fake news detection, we are provided with textual data of news, the propagation process of news, and interactive user profile data. Given total $N$ number of news as $\mathcal{N} = \{n_i\}_{i=1}^{N}$, propagation process of news $n_i$ is represented as a news propagation tree $p_i = (V_i, E_i)$, where $V_i$ is defined as the set of nodes and $X_i \in R^{|V_i| \times d}$ is their corresponding node features, $d$ is the feature dimension, nodes represent the source news and the users interacting with it. $E_i$ is defined as the set of edges, with each edge representing interactions like news-to-user reposts and user-to-user comments. The edges between nodes are represented using an adjacency matrix $A_i \in R^{|V_i| \times |V_i|}$.

A hypergraph can be denoted as $\mathcal{G} = (\mathcal{V}, \mathcal{E})$ where $\mathcal{V}$ denotes the node set and $\mathcal{E}$ denotes the hyperedge set. A hypergraph is an extension of graph where hyperedges allow for more than two nodes to be connected. Consequently, we use an incidence matrix $H \in R^{|\mathcal{V}| \times |\mathcal{E}|}$ to represent the structure of a hypergraph. Formally, for $v \in \mathcal{V}$ and $e \in \mathcal{E}$, we have:

$$H(v,e) = \begin{cases} 1, & if \quad v \in e; \\ 0, & otherwise, \end{cases} \quad (1)$$

### Problem Definition

Given a series of news $\mathcal{N} = \{n_i\}_{i=1}^{N}$, the corresponding propagation trees $\mathcal{P} = \{p_i\}_{i=1}^{N}$, and a hypergraph structure $\mathcal{G} = (\mathcal{V}, \mathcal{E})$ built from high-order relationships among news. We describe the fake news detection problem as a binary classification task, where our goal is to learn a function that maps news $n_i$ to predict the authenticity label $\hat{y}_i \in \{0,1\}$. In formulation,

$$[\{n_1, p_1\}, \{n_2, p_2\}, \ldots, \{n_N, p_N\}; \mathcal{G}] \rightarrow [\hat{y}_1, \hat{y}_2, \ldots, \hat{y}_N], \quad (2)$$

Besides, the notations used in this paper are shown in Table 1 for clarity.

## Method

This work proposes a novel model named DHy-MFND for fake news detection, which models news data from three distinct views. It utilizes a pre-trained language model to capture textual features of news, employs GNN to capture the propagation characteristics of individual news pieces, and utilizes HGNN to capture the high-order relationships among multiple news pieces. To enhance the hypergraph structure, we introduce a dynamic hypergraph structure learning module. This module dynamically optimizes the hypergraph structure by leveraging hyperedge embeddings and news embeddings from hypergraph convolution process. To extract features more relevant to authenticity, we incorporate contrastive learning between embeddings obtained from different views. This process involves minimizing the distance between news pieces with the same authenticity label while simultaneously maximizing the distance between those with differing authenticity labels. Finally, we integrate these blocks into a holistic multi-view fusion extraction module, which first extract textual features from each news, and then feeds data into both the propagation tree-level view and the hypergraph-level view parallelly. The framework of the model is illustrated in Figure 2.

| Notations | Descriptions |
|---|---|
| $\mathcal{N} = \{n_i\}_{i=1}^{N}$ | News pieces |
| $p_i = (V_i, E_i)$ | News propagation tree of $n_i$ |
| $\mathcal{G} = (\mathcal{V}, \mathcal{E})$ | News hypergraph |
| $A_i$ | Adjacency matrix of $p_i$ |
| $H, H_{re}$ | Incidence matrix, reconstructed incidence matrix |
| $\hat{y}_i, y_i$ | Ground truth and predict label of $n_i$ |
| $U^{(l)}, X_h^{(l)}$ | Hyperedge and node embedding of $l$-th HGNN |
| $X_{text}, X_{usr}$ | News textual embedding and user features |
| $X_{pro}$ | News propagation embedding |
| $X_{hg}$ | News hypergraph embedding |
| $X_{news}$ | Final news embedding |

Table 1: Notations and definition.

### Multi-view Fusion Fake News Detection Model

We model the news from three distinct views of varying scales, respectively.

**Text-level View.** For the textual data of news pieces, we follow the approach of UPFD (Dou et al. 2021) by utilizing a pre-trained language model to encode the textual information of the news. Given a series of news pieces $\mathcal{N} = \{n_i\}_{i=1}^{N}$, we employ the BERT pre-trained model to encode the news text and obtain the news textual features:

$$X_{text} = BERT(\mathcal{N}), \quad (3)$$

The $X_{text}$ is simultaneously fed into the other two views as the initial news feature.

**Propagation Tree-level View.** Given a series of news pieces $\mathcal{N} = \{n_i\}_{i=1}^{N}$ and their corresponding propagation trees $\mathcal{P} = \{p_i\}_{i=1}^{N}$, we utilize graph neural networks to encode each propagation tree. The construction of the propagation tree follows the strategy in (Dou et al. 2021; Shu et al. 2022), where the root node represents the source news and the other nodes represent the users interacting with it. The input of the propagation tree encoder includes the adjacency matrices of the propagation trees $\{A_i\}_{i=1}^{N}$, and the initial features of the nodes. The initial feature of the root node is the news text feature $X_{text}$, and the initial feature of the user nodes is the user preference $X_{usr}$. We extract the encoded root node feature from the propagation tree as the news propagation feature:

$$X_{pro} = Root\Big(GNN(X_{text}, X_{usr}, \{A_i\}_{i=1}^N)\Big), \quad (4)$$

here, $Root(\cdot)$ denotes the function for extracting the root node embedding from the propagation tree. $GNN(\cdot)$ represents the multi-layer graph neural network encoder, where we employ GraphSAGE as the backbone for the GNN.

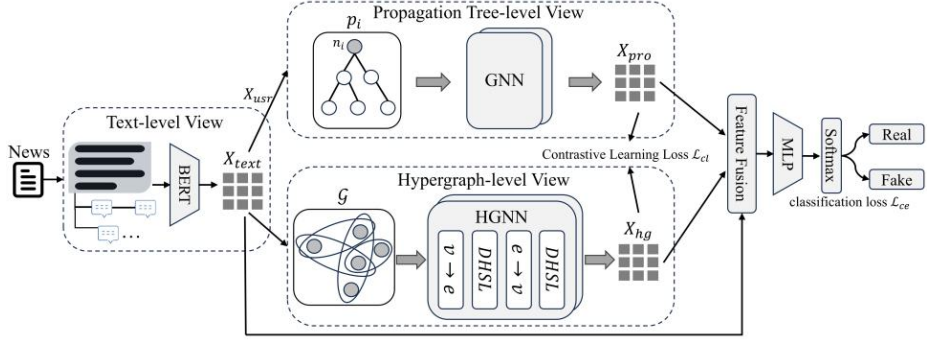

Figure 2: Overview of our proposed DHy-MFND.

**Hypergraph-level View.** Given a series of news pieces $\mathcal{N} = \{n_i\}_{i=1}^N$ and a hypergraph $\mathcal{G} = (\mathcal{V}, \mathcal{E})$ constructed based on the relationships between the news pieces, we employ a hypergraph neural network to encode the news hypergraph, capturing high-order relationships among news pieces. The predefined hypergraph structure we utilize follows the strategy in (Jeong et al. 2022) and is built based on three types of hyperedges: 1) connecting news pieces that are interacted with by the same user; 2) connecting news pieces published around the same time; 3) connecting news pieces that contain similar entities in their text content.

The input of the hypergraph neural network consists of the initial hypergraph incidence matrix $H$ and the initial news features $X_{text}$. The features of the hyperedges obtained during the hypergraph convolution process are denoted by $U$. We utilize the hypergraph neural network to encode the news hypergraph, capturing high-order relational features of the news:

$$X_{hg} = HGNN(X_{text}, H), \quad (5)$$

$HGNN(\cdot)$ represents a multi-layer hypergraph neural network encoder, where we employ hypergraph convolution based on attention mechanism (Bai et al. 2021). HGNNs can be viewed as a two-stage message passing process (Huang and Yang, 2021; Chien et al. 2021): update hyperedge features by aggregating all node features within the hyperedge, and then update node features by aggregating all hyperedge features that nodes belong to. Formally, the $l$-th layer of hypergraph convolution can be represented as:

$$U^{(l)} = f_{v2e}(X_h^{(l)}, H) = \sigma\Big((Att_{v2e} \odot H^T) X_h^{(l)} W_1\Big), \quad (6)$$
$$X_h^{(l+1)} = f_{e2v}(U^{(l)}, H) = \sigma\Big((Att_{e2v} \odot H) U^{(l)} W_2\Big), \quad (7)$$

$W_1, W_2 \in R^{d \times d}$ denotes trainable weight matrices, $Att_{v2e} \in R^{|\mathcal{E}| \times |\mathcal{V}|}$ and $Att_{e2v} \in R^{|\mathcal{V}| \times |\mathcal{E}|}$ are attention coefficient matrices from nodes to hyperedges and from hyperedges to nodes, respectively. $\sigma(\cdot)$ denotes a non-linear activation function, and $\odot$ indicates the Hadamard product. For a given node $v_i$ and its associated hyperedge $e_j$, the attentional score is:

$$(Att_{v2e})_{ji} = \frac{\exp\Big(\sigma\big(sim(u_j W, x_i W)\big)\Big)}{\sum_{k \in N_j} \exp\Big(\sigma\big(sim(u_j W, x_k W)\big)\Big)}, \quad (8)$$

$$(Att_{e2v})_{ij} = \frac{\exp\Big(\sigma\big(sim(x_i W, u_j W)\big)\Big)}{\sum_{k \in N_i} \exp\Big(\sigma(sim(x_i W, u_k W))\Big)}, \quad (9)$$

here, $N_j$ is the set of all nodes in hyperedge $e_j$, $N_i$ is the set of all hyperedges containing node $v_i$, and $W \in R^{d \times d}$ is a trainable weight matrix. $sim(\cdot)$ is a similarity function that computes the pairwise similarity between two vectors.

**News Embeddings Fusion.** We obtain news embeddings in three views $[X_{text}, X_{pro}, X_{hg}]$, and then adopt an attention layer to fuse the news representations adaptively, as follows:

$$X_{news} = Att_{fts}([X_{text}, X_{pro}, X_{hg}]), \quad (10)$$

$Att_{fts}(\cdot)$ denotes an attention layer with parameters $W_{Att} \in R^{1 \times 1 \times 3}$. Then the final news representation $X_{news}$ is input to a softmax function, and the two output neurons represent the probability of whether the prediction result is a fake news or true news:

$$\hat{y} = Softmax(X_{news}), \quad (11)$$

### Dynamic Hypergraph Structure Learning (DHSL)

The construction of hyperedges relies on strong assumptions, which can result in predefined hypergraph structures that may struggle to effectively capture the relationships necessary for downstream tasks or may introduce noise. To address this issue, we introduce hypergraph structure learning into hypergraph convolution, optimizing the hypergraph structure while learning news hypergraph embeddings. The process of dynamic hypergraph structure learning is illustrated in Figure 3.

To capture the complex relationships among news pieces, inspired by HSL (Cai et al. 2022) that utilizes similarity to capture potential connections, we leverage the news embeddings and hyperedge embeddings obtained from hypergraph convolution to compute similarity for capturing potential connections. Specifically, given the news embeddings $\tilde{X}_h$ and the hyperedge embeddings $\tilde{U}$, we calculate a similarity matrix $S \in R^{|\mathcal{V}| \times |\mathcal{E}|}$ using a weighted cosine similarity. Formally, the similarity is computed as:

$$s_{ij} = \cos(w_i \odot \tilde{x}_{h,i}, w_i \odot \tilde{u}_j), \quad (12)$$

here, $\{w_i\}_{i=1}^{n_{sim}}$ represents a weight vector, $\odot$ denotes the Hadamard product, $\tilde{x}_{h,i}$ and $\tilde{u}_j$ are the embeddings of the node $v_i$ and the hyperedge $e_j$ respectively. To reduce the expensive computation and noise, we filter out the nodes with the highest similarity to each hyperedge in $S$ based on a threshold ratio $0 \le p_{thd} \le 1$, resulting in the hypergraph structure $H_{ebd}$. Inspired by the approach of enhancing source node embeddings in news propagation trees (Bian et al. 2020), to strengthen the influence of source news, we similarly construct the hypergraph structure $H_{text}$.

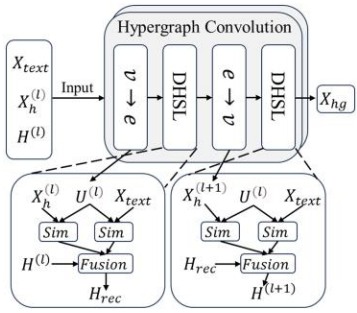

Figure 3: Dynamic hypergraph structure learning.

Therefore, we have obtained three hypergraph structures: $[H, H_{ebd}, H_{text}]$. We adopt a graph-level attention layer (Yun et al. 2019; Zhao et al. 2021) to fuse the hypergraphs adaptively, as follows:

$$H_{re} = Att_{\mathcal{G}}([H, H_{ebd}, H_{text}]), \quad (13)$$

here, $Att_{\mathcal{G}}(\cdot)$ denotes a graph-level attention layer with parameters $W_{Att} \in R^{1 \times 1 \times 3}$. Whenever the hyperedge embeddings or node embeddings are updated during the hypergraph convolution process, we can utilize them for hypergraph structure learning. The $l$-th layer of dynamic hypergraph structure learning conducted in the hypergraph convolution can be represented as:

$$U^{(l)} = f_{v2e}(X_h^{(l)}, H^{(l)}),$$
$$H_{re} = DHSL(X_h^{(l)}, X_{text}, U^{(l)}, H^{(l)});$$
$$X_h^{(l+1)} = f_{e2v}(U^{(l)}, H_{rec}),$$
$$H^{(l+1)} = DHSL(X_h^{(l)}, X_{text}, U^{(l)}, H_{rec}), \quad (14)$$

### Contrastive Learning

To enable the model to learn features more relevant to the authenticity, we introduce contrastive learning between news embeddings captured in the propagation tree-level view and hypergraph-level view. We construct a novel loss function as the optimization objective for fake news detection. Specifically, we classify pairs of news pieces with same authenticity labels as positive pairs, and those with differing labels as negative pairs. For an authentic news $n_i$, the aim is to enhance its cosine similarity with other authentic news while reducing its similarity with fake news. The final loss function comprises two parts: cross entropy loss and contrastive learning loss. We adopt the cross entropy to calculate the classification loss $\mathcal{L}_{ce}$, which can be obtained as follows:

$$\mathcal{L}_{ce} = -\frac{1}{N} \sum_{n_i \in \mathcal{N}} \sum_{c=1}^{C} \hat{y}_{i,c} \log(y_{i,c}), \quad (15)$$

where $y_{i,c}, \hat{y}_{i,c}$ denote the ground truth and the predicted probability of $n_i$ belonging to class $c \in \{1, \dots, C\}$, respectively.

We use $X_{pro}$ and $X_{hg}$ as feature representations of two contrastive views for contrastive learning. We apply InfoNCE (Oord et al. 2019) as the contrastive loss function to extract features more relevant to the authenticity. The contrastive learning loss can be formulated as follows:

$$\mathcal{L}_{cl} = -\frac{1}{N} \sum_{i=1}^{N} \log \left\{ \frac{1}{|K(i)|} \frac{\sum_{k \in K(i)} \exp(sim(x_{pro,i}, x_{hg,k})/\tau)}{\sum_{t \in T(i)} \exp(sim(x_{pro,i}, x_{hg,t})/\tau)} \right\}, \quad (16)$$

where the index $k$ corresponds to the positive sample with the same label as the sample at index $i$, and the index $t$ corresponds to the negative sample with the label different from the sample at index $i$. $K(i) = \{k: y_k = y_i\}$ is the set of indices of positive samples, and $T(i) = \{t: y_t \ne y_i\}$ is the set of indices of negative samples. $sim(\cdot)$ denotes the cosine similarity function, and $\tau$ is the temperature parameter.

The final loss for DHy-MFND can be described as follows:

$$\mathcal{L} = \mathcal{L}_{ce} + \lambda \mathcal{L}_{cl}, \quad (17)$$

Where, $\lambda$ is the balance parameter. The training algorithm of DHy-MFND is summarized in Algorithm 1.

---

**Algorithm 1: Training Algorithm of DHy-MFND**

**Input**: News pieces $\mathcal{N} = \{n_i\}_{i=1}^N$, News propagation trees $\mathcal{P} = \{p_i\}_{i=1}^N$, News hypergraph $\mathcal{G} = (\mathcal{V}, \mathcal{E})$, Labels $Y$
**Output**: News predicted authenticity label $[\hat{y}_1, \hat{y}_2, \dots, \hat{y}_N]$
1: Extract $X_{text}$ using Eq. (3)
2: **for** each epoch **do**
3:   Calculate $X_{pro}$ using Eq. (4)
4:   **for** $i = 1,2,\dots,L$ **do**
5:     Conduct node to hyperedge message passing
6:     Generate $H_{re}$ using Eq. (13)
7:     Conduct hyperedge to node message passing
8:     Generate $H_{re}$ using Eq. (13)
9:   **end for**
10:  Calculate $X_{hg}, X_{news}$ using Eq. (5) and Eq. (10)
11:  Calculate the final prediction label and classification loss
12:  Calculate contrastive loss $\mathcal{L}_{cl}$ between $X_{pro}$ and $X_{hg}$
13:  $\mathcal{L} = \mathcal{L}_{ce} + \lambda \mathcal{L}_{cl}$ and update the model parameters
14: **end for**

|  | PolitiFact | | Gossipcop | |
| --- | --- | --- | --- | --- |
|  | Acc (%) | F1 (%) | Acc (%) | F1 (%) |
| BERT | 71.04±3.46 | 71.03±2.75 | 85.76±1.21 | 85.75±2.17 |
| FinerFact | 91.48±1.89 | 91.48±1.89 | 86.57±0.49 | 86.44±0.49 |
| GNNCL | 65.79±8.96 | 65.02±9.46 | 94.98±0.80 | 94.94±0.80 |
| BiGCN | 74.16±3.57 | 74.16±3.57 | 88.04±0.48 | 87.95±0.49 |
| UniPF | 90.69±1.48 | 90.43±1.48 | 96.82±0.67 | 96.82±0.67 |
| GCNFN | 80.63±4.23 | 80.31±4.57 | 95.37±0.21 | 95.33±0.21 |
| UPFD-GCN | 80.27±4.35 | 80.16±4.41 | 95.55±0.63 | 95.51±0.64 |
| UPFD-GAT | 79.09±3.73 | 78.95±3.79 | 96.03±0.62 | 96.00±0.62 |
| UPFD-SAGE | 80.40±4.22 | 80.13±4.65 | 96.38±0.48 | 96.36±0.48 |
| HGNN | 79.96±4.89 | 79.28±5.16 | 93.38±0.49 | 93.38±0.49 |
| HGFND | 90.31±1.50 | 90.31±1.50 | 97.46±0.30 | 97.46±0.30 |
| DHy-MFND | **92.81±2.50** | **92.69±2.59** | **98.84±0.30** | **98.84±0.30** |

Table 3: Comparisons on fake news detection (%): Mean accuracy ± standard deviation. The best results are in bold.

## Experiment

### Experimental Setup

| Datasets | Graph | True | Fake | Nodes | Edges |
| --- | --- | --- | --- | --- | --- |
| Politifact | 314 | 157 | 157 | 41054 | 40740 |
| Gossipcop | 5464 | 2732 | 2732 | 314262 | 308798 |

Table 2: Datasets statistics.

**Datasets.** We use two fact-checking datasets, PolitiFact and Gossipcop, both sourced from FakeNewsNet (Shu et al. 2020). PolitiFact focuses on political news, while Gossipcop collects news related to entertainment. Through collection and preprocessing in UPFD (Dou et al. 2021), the dataset includes not only fact-checked source news but also user interactions with news and user profile information. The predefined hypergraph structure is derived from HGFND (Jeong et al. 2022).

**Evaluation Metrics.** We measure the performance of our proposed DHy-MFND and baseline methods for fake news detection using two widely-used metrics: Accuracy (Acc) and F1 score (F1).

**Baselines.** We will compare our proposed model against 9 fake news detection baselines, including 2 text mining methods: BERT (Devlin et al. 2019), FinerFact (Jin et al. 2022), 5 propagation tree methods: GNNCL (Han et al. 2020), BiGCN (Bian et al. 2020), GCNFN (Monti et al. 2019), UniPF (Wei et al., 2022), UPFD (Dou et al. 2021), and 2 hypergraph methods: HGNN (Feng et al. 2019) and HGFND (Jeong et al. 2022).

**Implementation Details.** The proposed model is implemented with PyTorch and is trained on an NVIDIA RTX GPU. We employ 2 layers of graph convolution as the propagation tree encoder, and employ 2 layers of hypergraph convolution as the hypergraph encoder. We split the dataset into training, validation, and testing sets in a 6:2:2 ratio. We utilize Adam as the optimizer, set the hidden layer dimension to 128, batch size to 64, learning rate to 0.001, dropout rate of 0.5, and run for 200 epochs. For the other parameters in the baselines, we apply the parameters suggested in each baseline. We use the pre-computed features of news and user's recent 200 tweet/retweet encoded by BERT (Devlin et al. 2019) for the proposed model, where the initial feature dimension is 768.

### The Performance of Our Method

The performance of DHy-MFND in comparison with baselines are listed in Table 3. To ensure the stability of the experimental results, all the models are repeated 20 times randomly and their means and standard deviations are reported. According to the results, we have several observations described as follows.

Firstly, the proposed DHy-MFND consistently outperforms in both Accuracy and F1 metrics on Politifact and Gossipcop, demonstrating the superiority of the proposed multi-view fake news detection method, dynamic hypergraph structure learning module, and contrastive learning approaching. Specifically, the improvement can be attributed to 3 aspects: (1) DHy-MFND learns news embeddings from three views across various scales, capturing a comprehensive range of news information at multiple levels. (2) The DHSL module optimizes the predefined hypergraph, capturing potential high-order relationships among news pieces. (3) Contrastive learning enables the model to capture embeddings that are more relevant to authenticity, thereby enhancing the quality of the learned embeddings.

Secondly, while FinerFact (Jin et al., 2022) exhibits notable performance on Politifact, it falls short in comparison to propagation-based approaches on Gossipcop. This highlights the efficacy of news propagation-based methods in capturing propagation information. Furthermore, hypergraph-based methods notably outperform propagation tree-based methods on Politifact more than on Gossipcop, underscoring the value of capturing news relationships in enhancing model performance, particularly in straightforward propagation process. Among these methods, DHy-MFND achieves the best performance, which shows the effectiveness of the proposed modules.

## Time Efficiency

| Datasets | Politifact | Gossipcop |
|---|---|---|
| DHy-MFND | 7.51s | 311.73s |

Table 4: Average training time on each epoch.

To evaluate the execution time of DHy-MFND, we record its training time on both datasets. Since the inputs of various methods differ, direct comparison of the runtime with other baseline methods may be unfair. Table 4 presents the average runtime per epoch of DHy-MFND on both datasets. The longer runtime on Gossipcop is a result of the higher number of news nodes in the hypergraph. Overall, DHy-MFND effectively learns news embeddings within an acceptable runtime.

## Ablation Studies

|  | Politifact | | Gossipcop | |
|---|---|---|---|---|
|  | Acc (%) | F1 (%) | Acc (%) | F1 (%) |
| w/o Text | 91.43 | 91.25 | 97.86 | 97.86 |
| w/o Pro | 82.34 | 82.22 | 92.41 | 92.41 |
| w/o HG | 83.90 | 83.77 | 97.44 | 97.43 |
| w/o CL | 88.82 | 88.76 | 97.46 | 97.47 |
| w/o DHSL | 91.48 | 91.39 | 97.38 | 97.37 |
| DHy-MFND | **92.81** | **92.69** | **98.84** | **98.84** |

Table 5: Comparisons of the DHy-MFND with its 5 variants on fake news detection. The best results are in bold.

We compare DHy-MFND with its 5 variants on both datasets to intuitively understand the effectiveness of each component. To evaluate the contributions of each view, we created several variants by removing specific views from the model, such as removing the text view (w/o Text), the propagation tree view (w/o Pro), and the hypergraph (w/o HG). It is important to note that even in variants without the text view, BERT-encoded news text features are still utilized as the initial input features for the other views. To assess the significance of contrastive learning, we created a model variant by excluding the InfoNCE loss from the final loss function (w/o CL). To evaluate the importance of dynamic hypergraph structure learning, we devised a model variant by removing the DHSL module from the model (w/o DHSL).

From the results in Table 5, we can draw the following conclusions: (1) DHy-MFND consistently outperforms both w/o CL and w/o DHSL, underscoring the effectiveness of DHSL and contrastive learning. (2) DHy-MFND consistently outperforms all variants that remove individual views, suggesting that each view contributes significantly to the model's performance. (3) w/o HG outperforms w/o Pro highlights the importance of propagation process data for fake news detection. Meanwhile, the notable performance of w/o Text can be attributed to the role of news text characteristics as the foundational representation for other views.

## Hyperparameter Analysis

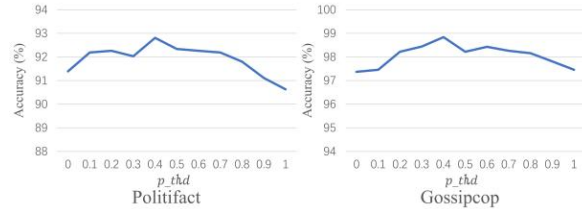

Figure 4: Parameter sensitivity analysis of $p_{thd}$.

In this section, we analyze the predefined hyperparameters in the model. The similarity threshold $p_{thd}$ used in dynamic hypergraph structure learning controls the number of nodes in generated similar hyperedges. Using a ratio instead of a specific value as the threshold allows for easy parameter adjustment and control over the appropriate number of nodes per hyperedge in the new hypergraph structure. For $0 \leq p_{thd} \leq 1$, we conduct repeated experiments with values in the range from 0 to 1 with an interval of 0.1. For $0 \leq p_{thd} \leq 1$, we conducted repeated experiments by incrementally selecting values at intervals of 0.1.

As illustrated in Figure 4, when $p_{thd} = 0$, the model aligns with a version of DHy-MFND lacking DHSL, corresponding with the results of w/o DHSL in the ablation studies. The observation indicates that a small number of potential connections is not sufficient to obtain informative news embeddings, whereas too many may introduce noise and weaken the complex relationships capturing. Overall, the visualization indicates that DHy-MFND is less sensitive to $p_{thd}$ in general and achieves high performance over a relatively wide range of values.

## Conclusion

In this paper, to address the existing shortcomings, we propose a novel multi-view fake news detection model. Initially, our method captures text features, propagation patterns, and high-order relationships of news across three views, to learn news features effectively without additional information. Subsequently, through inter-view contrastive learning, we improve feature quality for fake news detection. Furthermore, we introduce dynamic hypergraph structure learning to optimize the predefined hypergraph structure. On two real-world datasets for fake news detection, our model outperforms state-of-the-art methods.

# References

Kaliyar, R.K.; Goswami, A.; and Narang, P. 2021. FakeBERT: Fake news detection in social media with a BERT-based deep learning approach. *Multimedia tools and applications* 80(8): 11765-11788.

Rao, D.; Xin, M.; Jiang, Z.; and Li, Ran. 2021. STANKER: Stacking Network Based on Level-Grained Attention-Masked BERT for Rumor Detection on Social Media. In Proceedings of the 2021 Conference on Empirical Methods in Natural Language Processing, 3347-3363. doi.org/10.18653/v1/2021.emnlp-main.269.

Gong, S.; Sinnott, R.O.; Qi, J.; and Paris, C. 2023. Fake news detection through graph-based neural networks: A survey. arXiv:2307.12639.

Gao, Y.; Zhang, Z.; Lin, H.; Zhao, X.; Du, S.; and Zou, C. 2021. Hypergraph Learning: Methods and Practices. *IEEE Transactions on Pattern Analysis and Machine Intelligence* 44(5): 2548-2566. doi.org/10.1109/tpami.2020.3039374.

Yang, Y.; Huang, C.; Xia, L.; Liang, Y.; Yu, Y.; and Li, C. 2022. Multi-Behavior Hypergraph-Enhanced Transformer for Sequential Recommendation. In Proceedings of the 28th ACM SIGKDD Conference on Knowledge Discovery and Data Mining, 2263-2274. doi.org/10.1145/3534678.3539342.

Feng, Y.; You, H.; Zhang, Z.; Ji, R.; and Gao, Y. 2019. Hypergraph Neural Networks. In Proceedings of the AAAI Conference on Artificial Intelligence, 3558–3565. doi.org/10.1609/aaai.v33i01.33013558.

Jiang, J.; Wei, Y.; Feng, Y.; Cao, J.; and Gao, Y. 2019. Dynamic Hypergraph Neural Networks. In Proceedings of the Twenty-Eighth International Joint Conference on Artificial Intelligence, 2635-2641. doi.org/10.24963/ijcai.2019/366.

Wei, L.; Hu, D.; Lai, Y.; Zhou, W.; and Hu, S. 2022. A unified propagation forest-based framework for fake news detection. In Proceedings of the 29th International Conference on Computational Linguistics, 2769-2779.

Monti, F.; Frasca, F.; Eynard, D.; Mannion, D.; and Bronstein, M. M. 2019. Fake news detection on social media using geometric deep learning. arXiv:1902.06673.

Han, Y.; Karunasekera, S.; and Leckie, C. 2020. Graph neural networks with continual learning for fake news detection from social media. arXiv: 2007.03316.

Wu, H.; Li, N.; Zhang, J.; Chen, S.; Ng, M.K.; Long, J. 2024. Collaborative contrastive learning for hypergraph node classification. *Pattern Recognition* 146: 109995.

Ruggeri, N.; Contisciani, M.; Battiston, F.; and De Bacco, C. 2023. Community detection in large hypergraphs. arXiv:2301.11226.

Wang, C.; Wang, X.; Li, Z.; Chen, Z.; and Li, J. 2023. Hyconve: A novel embedding model for knowledge hypergraph link prediction with convolutional neural networks. In Proceedings of the ACM Web Conference, 188-198.

Jin, Y.; Wang, X.; Yang, R.; Sun, Y.; Wang, W.; Liao, H.; and Xie, X. 2022. Towards Fine-Grained Reasoning for Fake News Detection. In Proceedings of the AAAI Conference on Artificial Intelligence, 5746–5754. doi.org/10.1609/aaai.v36i5.20517.

Lu, Y. Ju.; and Li, C.T. 2020. GCAN: Graph-Aware Co-Attention Networks for Explainable Fake News Detection on Social Media. arXiv:2004.11648.

Bian, T.; Xiao, X.; Xu, T.; Zhao, P.; Huang, W.; Rong, Y.; and Huang, J. 2020. Rumor Detection on Social Media with Bi-Directional Graph Convolutional Networks. In Proceedings of the AAAI Conference on Artificial Intelligence, 549–556. doi.org/10.1609/aaai.v34i01.5393.

Dou, Y.; Shu, K.; Xia, C.; Yu, P.S.; and Sun, L. 2021. User Preference-Aware Fake News Detection. In Proceedings of the 44th International ACM SIGIR Conference on Research and Development in Information Retrieval, 2051-2055. doi.org/10.1145/3404835.3462990.

Su, X.; Yang, J.; Wu, J.; and Zhang, Y. 2022. Mining User-Aware Multi-Relations for Fake News Detection in Large Scale Online Social Networks. In Proceedings of the sixteenth ACM international conference on web search and data mining, 51-59. doi.org/10.1145/3539597.3570478.

Zhang, J.; Chen, Y.; Xiao, X.; Lu, R.; and Xia, S. 2021. Learnable hypergraph laplacian for hypergraph learning. arXiv:2106.06666.

Cai, D.; Song, M.; Sun, C.; Zhang, B.; Hong, S.; and Li, H. 2022. Hypergraph Structure Learning for Hypergraph Neural Networks. In IJCAI, 1923-1929.

He, Z.; Li, C.; Zhou, F.; and Yang, Y. 2021. Rumor Detection on Social Media with Event Augmentations. In Proceedings of the 44th International ACM SIGIR Conference on Research and Development in Information Retrieval, 2020-2024. doi.org/10.1145/3404835.3463001.

Sun, T.; Qian, Z.; Dong, S.; Li, P.; and Zhu, Q. 2022. Rumor Detection on Social Media with Graph Adversarial Contrastive Learning. In Proceedings of the ACM Web Conference, 2789-2797. doi.org/10.1145/3485447.3511999.

Oord, A.; Li, Y.; and Vinyals, O. 2018. Representation Learning with Contrastive Predictive Coding. arXiv:1807.03748.

Shu, K.; Mahudeswaran, D.; Wang, S.; and Liu, H. 2022. Hierarchical Propagation Networks for Fake News Detection: Investigation and Exploitation. In Proceedings of the International AAAI Conference on Web and Social Media, 626–637. doi.org/10.1609/icwsm.v14i1.7329.

Nguyen, V.H.; Sugiyama, K.; Nakov, P.; and Kan, M.Y. 2021. Fang: Leveraging social context for fake news detection using graph representation. In Proceedings of the 29th ACM international conference on information & knowledge management, 1165-1174. doi.org/10.1145/3340531.3412046.

Cui, J.; Kim, K.; Na, S.H.; and Shin, S. 2022. Meta-Path-Based Fake News Detection Leveraging Multi-Level Social Context Information. In Proceedings of the 31st ACM international conference on information & knowledge management, 325-334.

Yang, R.; Wang, X.; Jin, Y.; Li, C.; Lian, J.; and Xie, X. 2022. Reinforcement subgraph reasoning for fake news detection. In Proceedings of the 28th ACM SIGKDD Conference on Knowledge Discovery and Data Mining, 2253-2262. doi.org/10.1145/3534678.3539277.

Shu, K.; Mahudeswaran, D.; Wang, S.; Lee, D.; and Liu, H. 2020. FakeNewsNet: A Data Repository with News Content, Social Context, and Spatiotemporal Information for Studying Fake News on Social Media. *Big Data* 8(3): 171–188. doi.org/10.1089/big.2020.0062.

Jeong, U.; Ding, K.; Cheng, L.; Guo, R.; Shu, K.; and Liu, H. 2022 Nothing stands alone: Relational fake news detection with hypergraph neural networks. *IEEE International Conference on Big Data* 596-605.

Devlin, J.; Chang, M.W.; Lee, K.; Toutanova, K. 2018. Bert: Pre-training of deep bidirectional transformers for language understanding. arXiv:1810.04805.

Huang, J.; and Yang, J. 2021. Unignn: a unified framework for graph and hypergraph neural networks. arXiv:2105.00956.

Chien, E.; Pan, C.; Peng, J.; and Milenkovic, O. 2021. You are allset: A multiset function framework for hypergraph neural networks. arXiv:2106.13264.

Bai, S.; Zhang, F.; and Torr, P.H.S. 2021. Hypergraph Convolution and Hypergraph Attention. *Pattern Recognition*, 110, 107637. doi.org/10.1016/j.patcog.2020.107637.

Yun, S.; Jeong, M.; Kim, R.; Kang, J.; and Kim, H.J. 2019. Graph transformer networks. *Advances in neural information processing systems*, 32.

Zhao, J.; Wang, X.; Shi, C.; HU, B.; Song, G.; and Ye, Y. 2021. Heterogeneous graph structure learning for graph neural networks. In Proceedings of the AAAI conference on artificial intelligence., 35(5): 4697-4705.